%% file: paper.tex
\documentclass{article}

\PassOptionsToPackage{numbers, compress}{natbib}

\usepackage[utf8]{inputenc} 
\usepackage[T1]{fontenc}    
\usepackage{courier}
\usepackage{hyperref}       
\usepackage{url}            
\usepackage{booktabs}       
\usepackage{amsfonts}       
\usepackage{nicefrac}       
\usepackage{microtype}      
\usepackage{xcolor}         

\usepackage{multirow}
\usepackage{graphicx}
\usepackage{amsmath}
\usepackage{marvosym}
\definecolor{repoblue}{HTML}{0066CC}

\usepackage[main, final]{neurips_2026}

\hypersetup{
    colorlinks=true,
    linkcolor=blue,
    citecolor=blue,
    urlcolor=blue
}

\input{commands/commands}

\title{Controllable Road Marking Generation}

\author{%
  Zhiyu (Joey) Cai \quad Yufan Zhang \quad Ruichen Tan \quad Zengxiang Lei \quad Satish Ukkusuri\textsuperscript{\Letter} \\[0.5em]
  Purdue University, West Lafayette, IN 47906 \\
  \texttt{\{zjcai,zhan6037,tan479,lei67,sukkusur\}@purdue.edu}
}

\begin{document}

\maketitle

\vspace{-1em}
{\centering
  \hypersetup{hidelinks}
  \href{https://github.com/RU0804/CRMG}{%
    \textcolor{repoblue}{\texttt{https://github.com/RU0804/CRMG}}%
  }\par
}

\vspace{1em}

\begingroup
  \renewcommand{\thefootnote}{\Letter}
  \footnotetext{Corresponding author.}
\endgroup

\begin{abstract}
Lane and road markings provide critical guidance for vehicle navigation and multi-agent coordination, yet authoring them at scale remains a manual workflow that limits quantitative analysis and scenario testing. We introduce \emph{Controllable Road Marking Generation}, which synthesizes a missing center-region marking layout from a drivable-area mask, optional outer-ring markings, and a textual description. Our benchmark uses deterministic, metadata-derived prompts and three output channels: lane dividers, road dividers, and pedestrian crossings. We develop a conditional bird's-eye-view (BEV) pipeline that combines (i) a text-conditioned latent rectified-flow DiT trained with a topology-aware auxiliary loss, (ii) Gaussian-blurred training targets that provide smoother supervision for thin, sparse markings, and (iii) Structured Gaussian Render (SGR), a training-free post-process that recovers crisp divider geometry by extracting polylines, fitting cubic Bézier curves, and re-rendering them as anisotropic super-Gaussian primitives. On 4,597 Argoverse~2 test tiles, our system achieves Buffered F1 of 76.8 and clDice of 50.2, compared with 38.8 and 24.6 for an adapted mask-refinement baseline. When trained independently on the Waymo Open Dataset, it yields 88.0 Buffered F1 and 66.2 clDice. Component ablations show complementary connectivity gains from topology-aware supervision and SGR. Text-editing experiments reveal that stronger guidance improves edit success but also increases changes to non-target structures. We see this framework as a step toward simulation-ready road-marking variation, automated map completion, and early-stage infrastructure design exploration. 

\end{abstract}

\input{text/introduction}

\input{text/literature}

\input{text/problem}

\input{text/method}

\input{text/results}

\input{text/conclusion}

\newpage

\begin{ack}
\paragraph{Funding.}
The authors received no third-party funding or in-kind support
for this work during the 36 months before submission.

\paragraph{Competing interests.}
The authors declare no competing interests relevant to this work
during the 36 months before submission.
\end{ack}

\bibliographystyle{plainnat}
\bibliography{bib/references}


\input{text/appendix}

\clearpage






\end{document}

%% file: commands/commands.tex
\usepackage{amsmath}

\usepackage{color}

\usepackage{ifthen}
\usepackage{xcolor}

\newboolean{commentsOn}
\setboolean{commentsOn}{false}

\ifthenelse{\boolean{commentsOn}}{
\newcommand{\zhiyu}[1]{{\color{purple}{\bf ZC:} #1}}}
{\newcommand{\zhiyu}[1]{}}

\ifthenelse{\boolean{commentsOn}}{\newcommand{\yufan}[1]{{\color{red}{\bf YZ:} #1}}}
{\newcommand{\yufan}[1]{}}

\ifthenelse{\boolean{commentsOn}}{\newcommand{\zengxiang}[1]{{\color{orange}{\bf ZL:} #1}}}
{\newcommand{\zengxiang}[1]{}}

\ifthenelse{\boolean{commentsOn}}{\newcommand{\ruichen}[1]{{\color{brown}{\bf RT:} #1}}}
{\newcommand{\ruichen}[1]{}}

%% file: text/introduction.tex
\section{Introduction}

Lane and road markings are the visual syntax through which roads communicate traffic rules. They define lane boundaries, turning permissions, pedestrian priority, and other constraints that human drivers and automated vehicles use to interpret a scene~\cite{mosbock2016importance,babic2020road,biermeier2025road}. Their quality affects measurable behavior and safety outcomes~\cite{carlson2009benefits,gitelman2017examination}. For autonomous-driving research, road markings are also a compact machine-readable map layer: they constrain planning, provide simulation context, and determine marking-level edge cases for perception and control. 

Road-marking design remains a challenging workflow to scale. It requires civil-engineering expertise, regional guideline awareness, and substantial manual drafting or repair, which limits quantitative analysis and the exploration of design alternatives. Existing computational tools address adjacent parts of this workflow. Procedural generators and driving simulators instantiate road networks from hand-written templates~\cite{parish2001procedural,dosovitskiy2017carla}; scenario-generation methods diversify agents and trajectories on fixed maps~\cite{tan2021scenegen,pronovost2023scenario}; and HD-map or lane-marking models reconstruct observed maps from sensor, aerial, or partial-map evidence~\cite{liu2023vectormapnet,liao2022maptr,yuan2024streammapnet,qiao2023end,ruiz2025lane,monninger2025mapdiffusion}. This motivates a generative formulation of the road-marking layer itself: \emph{given a drivable-area geometry, optional surrounding markings, and a semantic request, what plausible marking layout could occupy the target region?}

We formulate this problem as \emph{Controllable Road Marking Generation}. Given a $100\,\mathrm{m}\times100\,\mathrm{m}$ bird's-eye-view (BEV) tile,
the task is to generate the center-region road-marking mask conditioned
on a drivable-area mask, an optional outer-ring observation of existing
markings, and a textual description. The output contains three channels: lane dividers, road dividers, and pedestrian crossings. In our benchmark,
descriptions are generated deterministically from tile metadata. A useful generator should place markings within feasible road geometry, preserve thin divider topology, remain consistent with surrounding context, and respond measurably to semantic edits.

The spatial and textual conditions play different roles in this task. Road geometry specifies the available road surface, while surrounding markings provide evidence about the configuration beyond the missing region. Text expresses the requested attributes of the completion. This separation supports two uses within the same formulation: recovering a layout consistent with an observed map, and examining how the generated layout changes when a textual attribute is edited. In the latter case, keeping the spatial inputs and random seed fixed allows changes in the target markings and unintended changes elsewhere to be assessed together.

This setting poses a technical challenge because road markings are sparse, high-frequency, topology-constrained structures. A one- or two-pixel displacement can break a dashed divider or merge nearby parallel dividers, while aggregate pixel losses may still look acceptable because background dominates the raster. Direct training on hard binary targets therefore tends to produce softened, fragmented, or merged markings. We address this with a two-stage blur-then-render strategy: Gaussian-blurred targets provide a stable continuous-valued training field, and \emph{Structured Gaussian Render} (SGR) recovers sharp lane- and road-divider geometry by extracting divider chains, fitting cubic B\'ezier curves, and re-rendering them with anisotropic super-Gaussian primitives.

This formulation could support offline map completion, simulation, and benchmark construction. Simulators and safety assessment pipelines gain map variation beyond fixed maps; urban design workflows gain rapid sketch-level alternatives; and controllable BEV map generation lets researchers test downstream systems under marking-level interventions~\cite{dosovitskiy2017carla,miao2018computational,tan2021scenegen,pronovost2023scenario}. We establish an Argoverse~2 benchmark, independently evaluate the same pipeline on Waymo, and report nuScenes results as an additional diagnostic. Our experiments assess structural completion fidelity, component contributions, and the trade-off between text-edit success and preservation of non-target structures.

Our contributions are as follows:
\begin{enumerate}
\item We introduce a conditional BEV synthesis task that generates rasterized road-marking layouts from drivable geometry, optional marking context, and template-derived natural language specifications. This enables controllable road-marking synthesis beyond observation-based extraction..
\item We construct a three-channel Argoverse~2 benchmark and extend evaluation to Waymo Open Dataset and a diagnostic nuScenes split. We evaluate both structural imputation and prompt-driven control with multiple metrics such as Buffered F1, clDice, prompt-edit success, false activation, and target channel change.
\item We develop a latent rectified-flow DiT with topology-aware supervision and a blur-and-render strategy, combining Gaussian-blurred training targets with deterministic curve-based rendering post-processing.
\item We evaluate against an adapted baseline and a learned CNN deblur ablation, showing strong structural gains and mask-parsed prompt-edit behavior across lane dividers, road dividers, and pedestrian crossings.
\end{enumerate}


%% file: text/literature.tex
\section{Related work}
\label{sec:related_work}

\providecommand{\checkmark}{\ensuremath{\surd}}
\providecommand{\rwcmark}{\textcolor{green!50!black}{\ensuremath{\checkmark}}}
\providecommand{\rwpmark}{\textcolor{orange!85!black}{\ensuremath{\triangle}}}
\providecommand{\rwxmark}{\textcolor{red!60!black}{\ensuremath{\times}}}

\begin{table}[t]
\centering
\footnotesize
\setlength{\tabcolsep}{3.5pt}
\renewcommand{\arraystretch}{0.94}
\caption{\textbf{Capability positioning by output representation, generative modeling, text conditioning, and thin-structure modeling. $\checkmark$ denotes direct support, $\triangle$ partial or adjacent capability, and $\times$ outside scope.} \rwcmark{} direct support; \rwpmark{} partial or adjacent capability; \rwxmark{} outside scope. Only capabilities are compared in this table.}
\label{tab:related_capability}
\begin{tabular}{@{}lccccc@{}}
\toprule
Method
& Raster
& Generative
& Layout
& Text
& Thin geometry \\
\midrule
RoadMark-cGAN~\cite{cira2026roadmark}
& \rwcmark & \rwcmark & \rwxmark & \rwxmark & \rwpmark \\
LSR-DM~\cite{ruiz2025lane}
& \rwcmark & \rwcmark & \rwxmark & \rwxmark & \rwpmark \\
MapDiffusion~\cite{monninger2025mapdiffusion}
& \rwxmark & \rwcmark & \rwpmark & \rwxmark & \rwpmark \\
LaneDiffusion~\cite{wang2025lanediffusion}
& \rwxmark & \rwcmark & \rwxmark & \rwxmark & \rwcmark \\
PolyDiffuse~\cite{chen2023polydiffuse}
& \rwxmark & \rwcmark & \rwpmark & \rwxmark & \rwpmark \\
\midrule
\textbf{Ours}
& \rwcmark & \rwcmark & \rwcmark & \rwcmark & \rwcmark \\
\bottomrule
\end{tabular}
\end{table}

\subsection{Road-marking and HD-map reconstruction}
The concept of using pavement markings to regulate traffic has developed over more than a century, originating from the first documented painted centerline applied in 1911 by Edward N. Hines in Wayne County, Michigan~\cite{mdot_national_firsts}. Since that major safety innovation, road markings have evolved into a highly standardized system that encodes lane boundaries, priority rules, and pedestrian crossing behavior, and their visibility and design affect driving and walking safety~\cite{mosbock2016importance,babic2020road,gitelman2017examination}. Most computational work, however, treats markings as structures to be recovered from observations~\cite{zhang2021deep}. Ego-view lane marking methods detect or parameterize visible dividers from camera images, including curve-based formulations such as B\'ezierLaneNet~\cite{feng2022rethinking}. RoadMark-cGAN~\cite{cira2026roadmark} is closer to the road-marking domain, but it maps aerial orthophotos to visible road-marking masks through image-to-image translation, again making the input image the source of geometry. HD-map reconstruction follows the same observation-driven paradigm at map scale: aerial methods extract lane-level street maps~\cite{he2022lane}, while online vector-map systems such as VectorMapNet, MapTR, StreamMapNet, and BeMapNet reconstruct BEV map elements from onboard sensor features~\cite{liu2023vectormapnet,liao2022maptr,yuan2024streammapnet,qiao2023end}. These works motivate our BEV representation and curve-aware decoding, but their objective is perception or map reconstruction under strong sensory evidence. In contrast, Controllable Road Marking Generation synthesizes the unobserved center-region marking layer from drivable geometry, optional surrounding markings, and a text request.

\subsection{Generative map and mask refinement}
Recent generative models for road and map structure remain primarily reconstructive or uncertainty-oriented. General diffusion inpainting methods such as RePaint~\cite{lugmayr2022repaint} show that denoising priors can fill masked regions, but do not encode road-map semantics or user controls. LSR-DM~\cite{ruiz2025lane}, our closest baseline, uses diffusion to refine aerial lane-segmentation masks before lane-graph extraction; LOID~\cite{agrawal2025loid} similarly uses generative inpainting to complete occluded lane segments from observations. MapDiffusion~\cite{monninger2025mapdiffusion} samples vectorized HD maps from BEV features to estimate uncertainty, and LaneDiffusion~\cite{wang2025lanediffusion} generates BEV centerline priors to improve centerline graph learning. SegRefiner~\cite{wang2023segrefiner} formulates coarse-to-fine mask cleanup as a discrete diffusion process. PolyDiffuse~\cite{chen2023polydiffuse} diffuses structured polygonal or polyline parameters. These works apply generative modeling to structured maps and masks, but they either refine an existing prediction, condition on sensor-derived features, or presuppose a parametric output space. Our setting instead asks a raster diffusion model to produce semantically specified road-marking layouts from text and geometric conditions, then decodes thin divider geometry deterministically at inference. 

MapDreamer~\cite{brandes2026mapdreamer} generates lane-level maps from aerial imagery, while ControlMap~\cite{farag2026controlmap} generates structured HD maps with spatial control. Our task instead completes a raster marking layer from drivable geometry, surrounding markings, and template-derived text.

\subsection{Thin-structure priors and structured rendering}
Road markings are sparse, elongated, and topology-sensitive, so our method is related to thin-structure segmentation and primitive rendering. Topology-aware delineation~\cite{Mosinska_2018_CVPR}, soft skeleton objectives such as clDice~\cite{shit2021cldice}, and recent skeleton-guided losses~\cite{liu2024enhancing} address the weak supervision signal of pixel losses on curvilinear structures; we use this lineage to motivate topology-aware training, not as a complete generative solution. At inference, our SGR module follows the line of work that represents curves with differentiable or splatted primitives. B\'ezier Splatting~\cite{liu2025bezier} and Curve-Aware Gaussian Splatting~\cite{gao2025curve} bind anisotropic Gaussian primitives to fitted curves, while differentiable vector and Gaussian image representations~\cite{li2020differentiable,ma2022towards,ying2025sketchsplat,huang20242d,zhang2024gaussianimage,zhou2024diffgs} cover the surrounding rendering context. Our contribution is to adapt this curve-bound Gaussian rendering idea as a deterministic post-process for controllable road-marking diffusion, where the primitive representation can preserve diverse road-line geometries such as dashed dividers, close parallel strokes, and double-yellow centerlines. This design targets the representation bottleneck of thin divider channels: it sharpens the diffusion output without retraining the generator, while keeping road-line instances separable when small raster displacements would otherwise blur, merge, or fragment them.

%% file: text/problem.tex
\section{Problem formulation}
\label{sec:problem}
\label{ssec:problem_setup}

\paragraph{Representation and controls.}
We represent a road-marking layout as a binary multi-channel BEV raster $d\in\{0,1\}^{H\times W\times K}$, where the $H\times W$ grid covers a fixed metric region (in our experiments $100\,\mathrm{m}\times100\,\mathrm{m}$ at $0.195\,\mathrm{m}/\mathrm{px}$) and each channel carries one categorical element. We instantiate $K{=}3$ for lane dividers, road dividers, and pedestrian crossings. The generation is conditioned on $\mathbf{C}=(\mathbf{s},\mathbf{p},\mathbf{m})$: the drivable-area mask $\mathbf{s}\in\{0,1\}^{H\times W}$ specifies where markings may appear; the optional \emph{outer-ring} observation $\mathbf{p}$ contains the rasterized markings in the boundary band surrounding the inner generation window; and the textual description $\mathbf{m}$ describes attributes such as lane count, divider style, and crosswalk presence. The evaluated descriptions use deterministic templates derived from metadata. The outer ring turns unconstrained local synthesis into context-aware completion: it anchors boundary continuity, lets generated dividers connect to existing markings, and prevents the center crop from being treated as an isolated road patch.

\paragraph{Problem objective.}
Given $\mathbf{C}$, the goal is to sample a plausible road-marking layout $d$ that is consistent with the conditioning context. Plausibility is defined operationally rather than as traffic-engineering certification: markings should lie within the drivable region, avoid contradictory class overlaps, remain coherent as thin road-marking structures, respect the optional outer-ring context, and reflect the semantic intent of $\mathbf{m}$. These conditions do not determine a unique target. The same drivable area may admit several reasonable lane-divider, road-divider, or crossing configurations, so controllable road marking generation is a conditional multimodal synthesis problem rather than deterministic reconstruction of a single layout. Two raster-level consistency conditions are
\begin{equation*}
(1-\mathbf{s})\odot d_{:,:,k}=\mathbf{0}\ \;\forall k,\qquad \sum_{k=1}^{K} d_{u,v,k}\le 1\ \;\forall (u,v).
\end{equation*}
These conditions describe desired properties of the generated layouts.

\paragraph{Learning formulation.}
We therefore model a conditional distribution $p_\theta(d\mid\mathbf{C})$ over condition-consistent layouts. Given a dataset $\{(d^{(i)},\mathbf{C}^{(i)})\}_{i=1}^N$, an ideal statistical objective is to maximize the conditional likelihood, equivalently minimize the negative log-likelihood (NLL),
\begin{equation*}
\theta^{\star}=\arg\min_{\theta}\;\Big\{-\tfrac{1}{N}\sum_{i=1}^N \log p_\theta\!\left(d^{(i)}\mid \mathbf{C}^{(i)}\right)\Big\},
\end{equation*}
In practice, we train a clean-latent rectified-flow predictor with reconstruction and topology-aware auxiliary losses rather than directly optimizing this likelihood. Sec.~\ref{sec:method} specifies the implemented training objective and inference-time rendering.

%% file: text/method.tex
\section{Methodology}
\label{sec:method}

\subsection{Overview}
\begin{figure}[t]
    \centering
    \includegraphics[width=\textwidth]{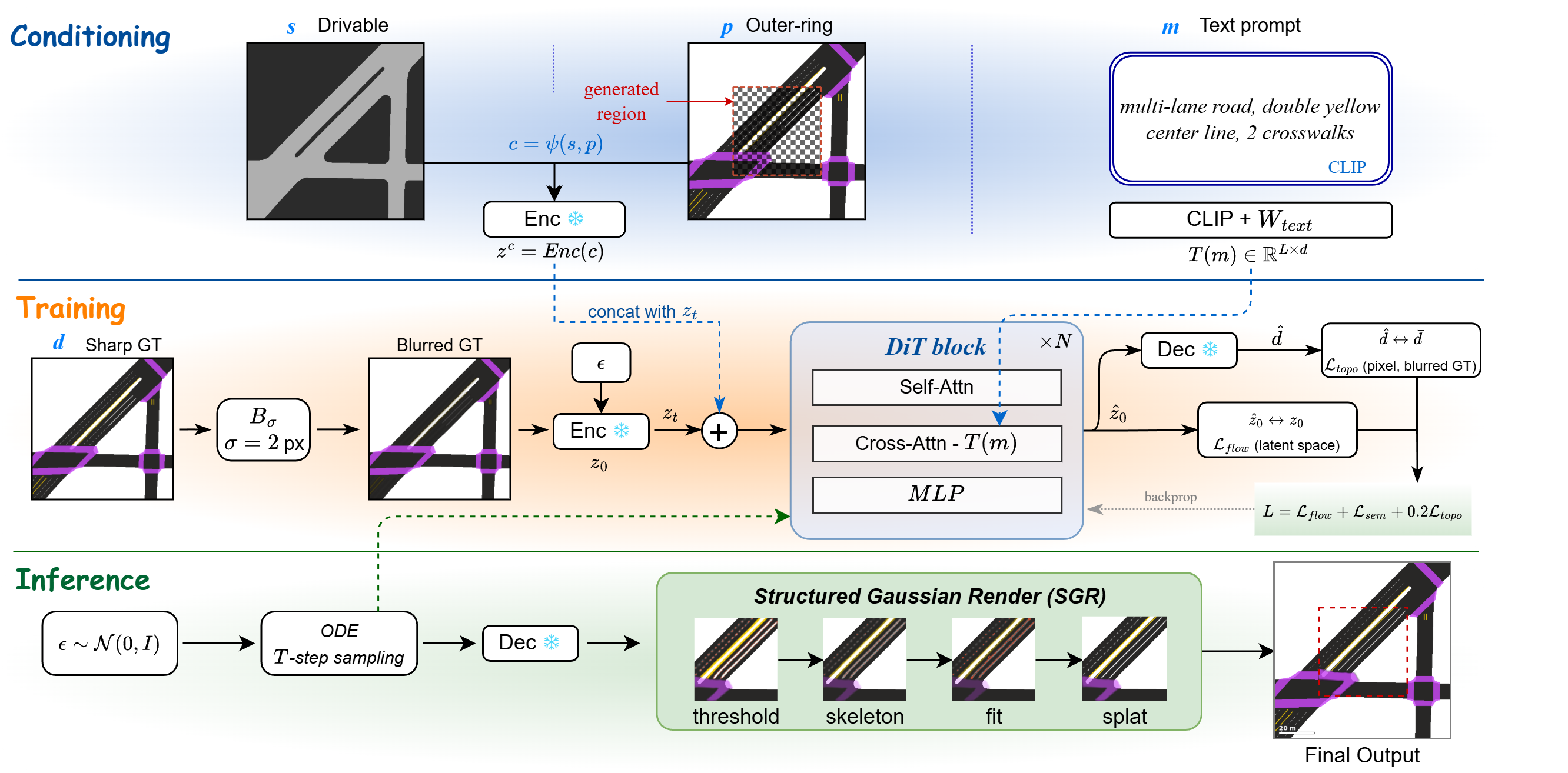}
    \caption{Overview of the proposed controllable road marking generation pipeline.}
    \label{fig:pipeline}
\end{figure}

The full pipeline is shown in Fig.~\ref{fig:pipeline}. We train a latent rectified-flow generator for the three-channel BEV marking raster $d$ conditioned on drivable area~$\mathbf{s}$, outer-ring markings~$\mathbf{p}$, and text~$\mathbf{m}$. The model produces a raw soft raster $\tilde{d}$; training blurs all center-window targets before autoencoder encoding, and inference sharpens the two divider channels with training-free Structured Gaussian Render (SGR). A lightweight CNN deblur module appears only as the learned-deblur ablation in Sec.~\ref{sec:experiments}.

The conditions are asymmetric: the drivable mask and outer-ring hint form a raster condition with no center target markings, while CLIP text tokens enter through gated cross-attention. Road geometry provides spatial context, and text provides semantic attributes. Lane count is used as descriptive context; controllability is evaluated for crosswalk presence and divider-style edits.

\subsection{Latent rectified flow with topology-aware loss}
\label{ssec:loss}

We use the rectified-flow straight-line interpolation of Liu et al.~\cite{liu2022flow}, parameterized as clean-latent prediction, in the latent space of a frozen pretrained autoencoder as in latent diffusion models~\cite{rombach2022high}. 
Let $\mathcal{B}_\sigma$ denote channel-wise Gaussian blurring of the center-window target with standard deviation $\sigma$, and define $\bar{d}=\mathcal{B}_\sigma(d)$ and $\mathbf{z}_0=\mathrm{Enc}(\bar{d})$.
Let $\psi$ construct a spatially aligned conditioning raster from the drivable-area mask $\mathbf{s}$ and observed outer-ring markings $\mathbf{p}$: the center contains drivable-area information only, while observed marking pixels are incorporated outside the center.
We form $\mathbf{c}=\psi(\mathbf{s},\mathbf{p})$ and encode it as $\mathbf{z}^{c}=\mathrm{Enc}(\mathbf{c})$.
With $\boldsymbol{\epsilon}\sim\mathcal{N}(0,\mathbf{I})$ and $t\sim\mathcal{U}(0,1)$, the interpolation, target velocity, and $x_0$-prediction parameterization are
\begin{equation*}
\mathbf{z}_t=(1-t)\mathbf{z}_0+t\boldsymbol{\epsilon},\qquad
\mathbf{v}^\star=\boldsymbol{\epsilon}-\mathbf{z}_0,\qquad
\hat{\mathbf{z}}_0=f_\theta(\mathbf{z}_t,t,\mathbf{z}^{c},\mathbf{T}),\qquad
\hat{\mathbf{v}}_\theta=\frac{\mathbf{z}_t-\hat{\mathbf{z}}_0}{\max(t,\tau)} ,
\end{equation*}
where $\mathbf{T}$ denotes the text-token sequence from Sec.~\ref{ssec:conditioning} and $\tau$ prevents division by zero near $t{=}0$. The reported model uses uniform timestep sampling.

The network is a DiT-style latent Transformer~\cite{peebles2023scalable}: it concatenates $\mathbf{z}_t$ and $\mathbf{z}^c$, patchifies the result, adds positional and timestep embeddings, and projects tokens back to $\hat{\mathbf{z}}_0$. Text enters through gated cross-attention; the outer ring enters through the condition latent.

\paragraph{Flow-matching and semantic reconstruction losses.}
The primary loss is a re-weighted squared error on the predicted clean latent, restricted to the central generation window:
\begin{equation*}
\mathcal{L}_{\text{flow}}=
\mathbb{E}\!\left[
\mathbf{M}_{\text{center}}\odot \mathbf{W}_{\text{mark}}\odot
(\hat{\mathbf{z}}_0-\mathbf{z}_0)^2
\right],
\qquad
\mathbf{W}_{\text{mark}}=\mathbf{1}+\alpha\,
\Big(A_{\downarrow}\big(\max_{k\leq 3}\bar{d}_k\big)\Big)^{\rho}.
\end{equation*}
Here $\mathbf{M}_{\text{center}}$ excludes the observed ring, $A_{\downarrow}$ maps marking support to latent resolution, and $\mathbf{W}_{\text{mark}}$ is a static foreground-weighted MSE analogue of class-imbalance reweighting~\cite{lin2017focal}.

Because pretrained autoencoder latents are not semantic output channels, we also supervise the decoded prediction in pixel space:
\begin{equation*}
\mathcal{L}_{\text{sem}}=
\mathbb{E}\!\left[
\mathbf{M}_{\text{center}}^{\text{pix}}\odot
\sum_{k=1}^{3}\beta_k(\hat{d}_k-\bar{d}_k)^2
\right],
\qquad
\hat{d}=\mathrm{Dec}(\hat{\mathbf{z}}_0).
\end{equation*}
The reported run uses $\boldsymbol{\beta}=(3.0,2.0,1.5)$ for lane dividers, road dividers, and pedestrian crossings, matching their relative sparsity and structural importance.

\paragraph{Topology-aware auxiliary loss.}
Pixel-wise reconstruction can be locally accurate while losing thin-structure connectivity~\cite{Mosinska_2018_CVPR}, so we add a clDice-style soft skeleton term~\cite{shit2021cldice} on the decoded clean prediction. Let $S_K(\cdot)$ be the differentiable skeletonizer, with $K{=}3$ in reported experiments. We use $C_{\mathrm{topo}}=3$ decoded channels and compute the following quantities per sample, with sums taken jointly over these channels and the full spatial tile:
\begin{equation*}
T_{p}=\frac{\sum S_K(\hat{d})\odot \bar{d}+\varepsilon}{\sum S_K(\hat{d})+\varepsilon},\qquad
T_{s}=\frac{\sum S_K(\bar{d})\odot \hat{d}+\varepsilon}{\sum S_K(\bar{d})+\varepsilon},\qquad
\ell_{\text{topo}}=1-\frac{2T_{p}T_{s}}{T_{p}+T_{s}+\varepsilon}.
\end{equation*}
We weight this term by $w(t)=\max(1-t,0.2)$ to emphasize low-noise predictions and cap saturated per-sample losses. The full objective is
\begin{equation*}
\mathcal{L}=\mathcal{L}_{\text{flow}}+\mathcal{L}_{\text{sem}}
+\lambda_{\text{topo}}\,\mathbb{E}\!\left[w(t)\,\min(\ell_{\text{topo}},0.95)\right],
\end{equation*}
with $\lambda_{\text{topo}}=0.2$. This adapts topology-aware and skeleton-based segmentation losses~\cite{Mosinska_2018_CVPR,shit2021cldice,liu2024enhancing} to decoded outputs of a latent generator.

\subsection{Conditioning inputs}
\label{ssec:conditioning}

\paragraph{Spatial condition.}
The spatial condition is $\mathbf{c}=\psi(\mathbf{s},\mathbf{p})$. The operator $\psi$ copies $\mathbf{p}$ only outside the center generation window, replaces condition pixels where an outer-ring marking exists, and leaves the center conditioned by drivable area alone. During training, $20\%$ of samples use a random block-wise keep ratio in $[0,1]$ over marking-bearing outer-ring cells; the rest keep the full outer ring, matching inference. Encoding $(\mathbf{s},\mathbf{p})$ exposes road width, curvature, intersection shape, and boundary-continuation cues through a joint condition latent concatenated with the noisy target latent.

This is a conditional-inpainting formulation: the center lacks markings, but its drivable-area geometry is still observed. A masked marking raster alone would discard that central geometric evidence. By constructing $\mathbf{c}=\psi(\mathbf{s},\mathbf{p})$, we retain both the observed road surface in the center and marking-continuation cues at the boundary in one aligned raster. The encoded condition therefore supplies information about possible line placement beyond a binary indication of where markings are allowed. Text can then influence semantic attributes within this geometric context.

All reported models use a fixed $512\times512$ raster covering
$100\,\mathrm{m}\times100\,\mathrm{m}$, with a $256\times256$
generation window. Arbitrary-window inference is not evaluated.

\paragraph{Text branch.}
The prompt~$\mathbf{m}$ is embedded with a frozen CLIP text encoder~\cite{radford2021learning} and projected to width $d_{\text{model}}$ by $W_{\text{text}}$, giving $\mathbf{T}=W_{\text{text}}\,E_{\text{clip}}(\mathbf{m})\in\mathbb{R}^{L\times d_{\text{model}}}$ with $L\le 77$. Prompts come from tile metadata and describe lane count, road direction, divider style, and crosswalk presence; the template is in Appendix~\ref{app:semantic_prompt}. We drop the prompt with probability $0.1$ during training for classifier-free guidance~\cite{ho2022classifier}. The CLIP encoder stays frozen; $W_{\text{text}}$, the text cross-attention modules, and the generator are trained jointly.

\paragraph{Cross-attention injection.}
Text enters through zero-initialized gated residual cross-attention.
The gate learns the text-conditioning strength; its cross-attention
structure follows Flamingo~\cite{alayrac2022flamingo}, while sharing
GLIGEN's residual-gating principle~\cite{li2023gligen}.

\subsection{Gaussian blur and inference-time Structured Gaussian Render}
\label{ssec:blur_deblur}

Thin binary markings are ill-conditioned targets for latent flow prediction: sparse one- or two-pixel strokes contribute little to pixel/latent MSE, yet small shifts can sever or merge topology~\cite{Mosinska_2018_CVPR,shit2021cldice}. We train on center-window targets blurred channel-wise with $\mathcal{B}_\sigma$, replacing each binary marking with a soft Gaussian profile. The reported system uses $\sigma{=}2$\,px, chosen from the pilot comparison in Appendix~\ref{app:gaussian_blur}; App.~\ref{app:components} evaluates the contributions of topology-aware supervision and SGR.

\begin{figure*}[t]
    \centering
    \includegraphics[width=\textwidth]{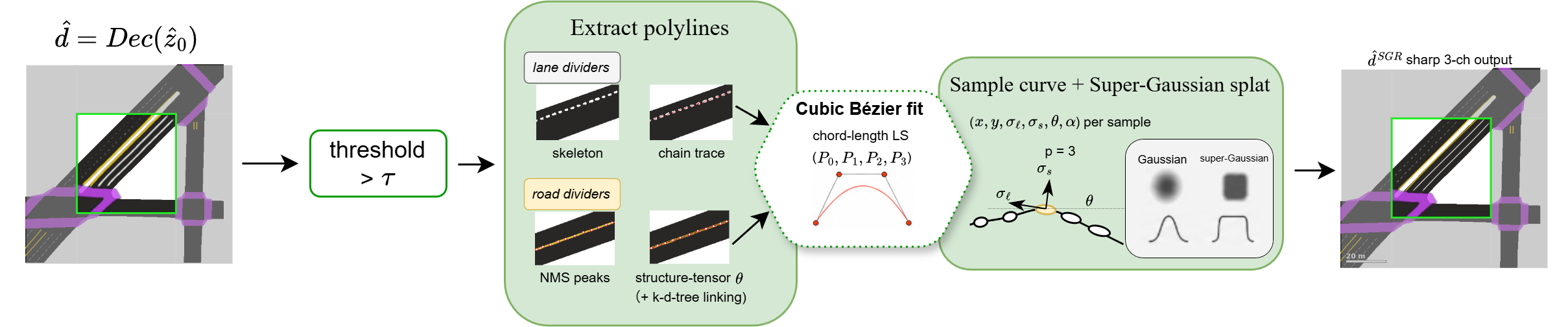}
    \caption{
    \textbf{Structured Gaussian Render (SGR).}
    Starting from the raw soft raster $\tilde{d}$, SGR thresholds the divider channels, extracts lane- and road-divider polylines, fits cubic B\'ezier curves, and re-renders sampled curve primitives with anisotropic super-Gaussian splats. The pedestrian-crossing channel is passed through as a raster mask.
    }
    \label{fig:sgr_pipeline}
\end{figure*}

At inference, the raw output $\tilde{d}$ inherits soft edges. As shown in Fig.~\ref{fig:sgr_pipeline}, we sharpen the two divider channels with \emph{Structured Gaussian Render} (SGR), a deterministic raster-to-curve-to-raster post-process. SGR shares the idea of rendering curve-tied Gaussian primitives with B\'ezier Splatting~\cite{liu2025bezier} and Curve-Aware Gaussian Splatting~\cite{gao2025curve}, but is a fixed mask post-process; the area-shaped crossing channel is passed through.

\paragraph{Per-channel curve extraction.}
For lane dividers, we threshold $\tilde{d}$, skeletonize the mask, remove junctions, and trace ordered polylines. For road dividers, we extract non-maximum-suppression peaks, estimate local orientation with structure-tensor/Sobel cues, link peaks with an angular gate, and split high-confidence plateaus with watershed. Each chain is fit by chord-length-parameterized least squares to a cubic B\'ezier curve.

\paragraph{Sampling, splatting, and composition.}
We sample each B\'ezier curve at channel-specific spacing, approximating uniform arc-length density without exact reparameterization. Each sample becomes an anisotropic super-Gaussian primitive $(x,y,\sigma_\ell,\sigma_s,\theta,\alpha)$, where $\theta$ is the curve tangent, $\sigma_s$ is a channel stroke-width prior, $\sigma_\ell$ follows sample spacing, and $\alpha$ is read from the soft-mask intensity. The order-$p$ kernel and per-channel alpha composition are
\begin{equation*}
g_i(u,v)=\exp\!\left(
-\frac{1}{2}
\left[
\left(\frac{u_i'}{\sigma_{\ell,i}}\right)^2+
\left(\frac{v_i'}{\sigma_{s,i}}\right)^2
\right]^{p/2}
\right),
\qquad
\hat{d}^{\mathrm{SGR}}_{u,v,k}=1-\!\prod_{i\in\mathcal{G}_k}\!\big(1-\alpha_i\,g_i(u,v)\big),
\end{equation*}
where $(u_i',v_i')$ is the displacement from primitive center $(x_i,y_i)$ after rotation by $-\theta_i$. We set $p{=}3$ for a flatter core and sharper falloff than a plain Gaussian, and use multiplicative alpha composition to keep composed intensities within $[0,1]$ without additive clipping.

SGR has no learned parameters; thresholds, NMS/linking parameters, stroke-width priors, target spacing, and super-Gaussian order are fixed on validation. It operates per channel without transferring probability mass across classes, and only the two divider channels are re-rendered. As a learned-deblur reference, we train a small residual CNN with an L1-plus-Dice objective, reported as \emph{Ours w/ CNN deblur} in Sec.~\ref{sec:experiments}.

The blur-and-render design assigns distinct roles to learning and geometric refinement. Blurred supervision gives thin markings a wider training signal, while the generator learns their placement and response to the conditioning inputs. SGR subsequently recovers sharp divider geometry from those predictions. It does not receive a text prompt or independently decide which semantic edit to make. This separation also informs the component analysis: generator-only outputs are skeletonized for width-normalized evaluation, without curve fitting or re-rendering, so that their broader pixel support does not by itself favor them over SGR outputs (App.~\ref{app:components}).

%% file: text/results.tex
\providecommand{\tbd}{\textsc{TBD}}
\providecommand{\notapp}{--}

\section{Evaluation and results}
\label{sec:experiments}

Evaluating road-marking generation presents two challenges.
First, prior work in this domain targets extraction, not design, so we cannot inherit a benchmark from earlier studies. Second, the design itself is multimodal: multiple plausible marking layouts may be compatible with a given drivable area, and a single ground truth therefore under-specifies success. We evaluate along two complementary axes: \emph{imputation accuracy} against an established map-refinement baseline using structural metrics on a held-out split (Sec.~\ref{ssec:main}), and \emph{conditional control}, measured by mask-parsed success and false-activation rates under single-attribute text edits with the spatial condition held fixed (Sec.~\ref{ssec:control}). The observed outer-ring markings constrain boundary continuity and the surrounding configuration, making single-reference metrics informative for matched completion.

\subsection{Experimental settings}
\label{ssec:experimental_settings}

\subsubsection{Data and preprocessing}
\label{ssec:data}

Our primary benchmark is Argoverse~2~\cite{wilson2023argoverse}, using local HD vector maps from six U.S. cities with lane geometry, boundary types, drivable areas, and pedestrian crossings. We follow the public split; rasterization, tile cropping, and the $100\,\mathrm{m}\times100\,\mathrm{m}\to 512{\times}512$ projection follow Sec.~\ref{ssec:problem_setup}. Metrics are computed only on the central $256{\times}256$ window, while the outer $128$\,px ring is conditioning context~$\mathbf{p}$.

For each tile, the spatial condition~$\mathbf{s}$ is the intersection of the drivable area and road-segment region, and the target~$\mathbf{d}$ has three channels: lane dividers, road dividers, and pedestrian crossings. Baselines use the same splits and preprocessing. We additionally evaluate independently trained models on the Waymo Open Dataset~\cite{sun2020waymo} and a diagnostic nuScenes-SG split~\cite{caesar2020nuscenes}; the corresponding results are reported in Tab.~\ref{tab:cross_dataset_structural_3ch}.

\subsubsection{Baselines}
\label{ssec:baselines}

Our primary baseline adapts LSR-DM~\cite{ruiz2025lane}, which refines aerial lane-segmentation masks with conditional diffusion before lane-graph extraction. We remove the unavailable aerial/CNN starting mask, provide the same drivable mask~$\mathbf{s}$ and outer ring~$\mathbf{p}$, predict the missing center markings, and retrain on our split. This matches spatial evidence, although the systems also differ in text conditioning, architecture, and post-processing; details are in App.~\ref{app:baseline_adaptation}. For matched completion, text prompts are constructed from ground-truth map annotations, including center-region metadata and crossing-mask information. Our model therefore receives semantic specifications in addition to the spatial inputs shared with the adapted baseline.

\subsubsection{Metrics}
\label{ssec:metrics}
To evaluate geometric fidelity and connectivity-sensitive structure, we use two per-channel structural metrics on the target area.

\paragraph{Centerline Dice (clDice):}
The clDice~\cite{shit2021cldice} compares each mask skeleton with the filled area of the other mask, emphasizing centerline continuity over small width changes. Let $P$ and $G$ denote binary prediction and ground truth, and let $S(\cdot)$ be morphological skeletonization:

\begin{equation*}
t_p = \frac{|S(P)\cap G|}{|S(P)|},\qquad t_s = \frac{|S(G)\cap P|}{|S(G)|},\qquad \mathrm{clDice}(P,G) = \frac{2t_p t_s}{t_p+t_s}.
\end{equation*}

\paragraph{Buffered Pixel F1 Score:}
Buffered Pixel F1 adds a small spatial tolerance before counting matches, inspired by buffer-based road-extraction evaluation and tolerance-based segmentation measures~\cite{heipke1997evaluation,csurka2013good}. With $\mathcal{D}_r(\cdot)$ denoting dilation by radius~$r$,

\begin{equation*}
P_r = \frac{|P\cap \mathcal{D}_r(G)|}{|P|},\qquad R_r = \frac{|G\cap \mathcal{D}_r(P)|}{|G|},\qquad \mathrm{F1}_{\mathrm{buf}}(r) = \frac{2P_rR_r}{P_r+R_r}.
\end{equation*}

We use $r=3$ pixels ($0.59\,\mathrm{m}$ at $0.195\,\mathrm{m}/\mathrm{px}$), allowing small rasterization or annotation offsets while still penalizing missing or misplaced markings.
Together, clDice penalizes broken or shifted marking centerlines, while Buffered F1 captures area overlap under a limited geometric tolerance.

For each channel, scores are averaged over test tiles with positive ground-truth support, and Overall is the unweighted mean of the three channel scores. An empty prediction on a GT-positive tile receives zero Buffered F1 and zero clDice.

\subsection{Comparison with baseline for imputation}
\label{ssec:main}

\begin{table*}[t]
\centering
\small
\setlength{\tabcolsep}{3.5pt}
\caption{
Structural metrics on $4{,}597$ AV2 test tiles, evaluated on the generated center $256{\times}256$ region. Per-channel scores use GT-positive tiles; Overall is the unweighted mean of the three per-channel scores. Values are percentages; bold/underline mark best/second-best.
}
\label{tab:main_structural_3ch}
\begin{tabular}{lcccccccc}
\toprule
\multirow{2}{*}{Method}
& \multicolumn{2}{c}{Overall}
& \multicolumn{2}{c}{Lane Div.}
& \multicolumn{2}{c}{Road Div.}
& \multicolumn{2}{c}{Ped. Cross.} \\
\cmidrule(lr){2-3}\cmidrule(lr){4-5}\cmidrule(lr){6-7}\cmidrule(lr){8-9}
& F1$\uparrow$ & clDice$\uparrow$
& F1$\uparrow$ & clDice$\uparrow$
& F1$\uparrow$ & clDice$\uparrow$
& F1$\uparrow$ & clDice$\uparrow$ \\
\midrule
LSR-DM\textsuperscript{\dag}~\cite{ruiz2025lane} & 38.8 & 24.6 & 37.3 & 15.6 & 43.6 & 27.9 & 35.4 & 30.4 \\
Ours w/ CNN deblur   & \underline{62.7} & \underline{46.3} & \underline{50.2} & \underline{18.5} & \underline{64.9} & \underline{37.3} & \underline{73.0} & \underline{83.1} \\
Ours w/ SGR deblur
& \textbf{76.8} & \textbf{50.2}
& \textbf{74.0} & \textbf{28.4}
& \textbf{66.6} & \textbf{38.0}
& \textbf{89.7} & \textbf{84.1} \\
\bottomrule
\end{tabular}
\par\smallskip
\footnotesize\textsuperscript{\dag}LSR-DM is adapted to our center-completion task; see App.~\ref{app:baseline_adaptation} for deviations.
\end{table*}

Our full system achieves 76.8 Buffered F1 and 50.2 clDice on AV2, compared with 38.8/24.6 for the adapted LSR-DM baseline and 62.7/46.3 for the CNN-deblur variant (Tab.~\ref{tab:main_structural_3ch}). Independently training the same pipeline on Waymo yields 88.0/66.2; nuScenes-SG is reported as a diagnostic because of its small, imbalanced split and incomplete drivable-area annotations (Tab.~\ref{tab:cross_dataset_structural_3ch}).
The topology/SGR ablation achieves the highest clDice with both components enabled, although Buffered F1 is not uniformly highest (Tab.~\ref{tab:topology_sgr}).

\begin{figure}[t]
    \centering
    \includegraphics[width=\textwidth]{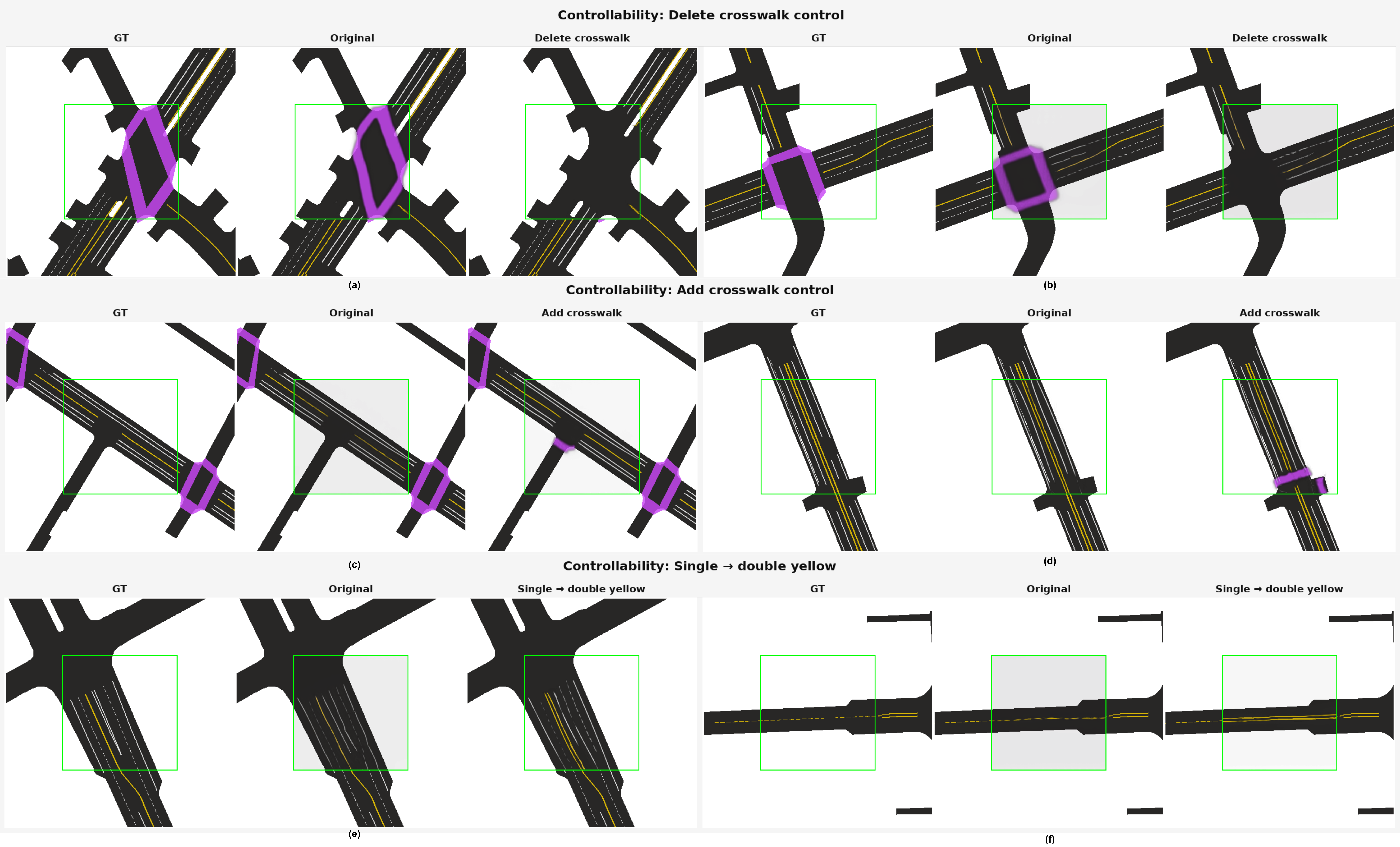}
    \caption{Qualitative text control. Each triplet shows the ground truth, the original-prompt generation, and the edited-prompt generation. Green boxes mark the generated center region.}
    \label{fig:control_viz}
\end{figure}

\subsection{Conditional control}
\label{ssec:control}

We next test whether text edits steer the generated layout while $(\mathbf{s},\mathbf{p})$ and the random seed remain fixed (Fig.~\ref{fig:control_viz}). For each tile, we edit one clause of $\mathbf{m}$ and parse the generated center mask. The protocol uses three \emph{mask-verifiable} edits (add crosswalk, delete crosswalk, and single$\to$double yellow), scored only on eligible tiles where the edit is non-trivial. Success measures the target-channel criterion, while false activation and \emph{fa-MSE} measure non-target side effects. To isolate text control from deblurring, parsing uses raw diffusion samples with GT blurred to training sharpness; templates and thresholds are in App.~\ref{app:semantic_prompt}.

\begin{table}[t]
\centering
\small
\setlength{\tabcolsep}{4pt}
\caption{
Mask-parsed controllability for single-clause prompt edits on the AV2 test split. Each family is evaluated only on eligible non-trivial tiles ($N_{\mathrm{elig}}$), with $(\mathbf{s},\mathbf{p})$ and seed fixed between baseline and edited prompts. Success and False activation are percentages; \emph{fa-MSE} is the mean squared change in non-target channel measures normalized by their on-thresholds, and $\Delta_{\mathrm{tgt}}$ is direction-aligned target change, with positive values indicating change in the requested direction. Protocol details are in App.~\ref{app:semantic_prompt}. Single-to-double yellow success uses a skeleton-length proxy rather than explicit verification of two parallel divider strokes. Target change is measured as crossing area in pixels for crosswalk edits and skeleton length in pixels for divider edits.
}
\label{tab:control_3ch}
\begin{tabular}{lccccc}
\toprule
Control family & $N_{\mathrm{elig}}$ & Success$\uparrow$ & False act.$\downarrow$ & fa-MSE$\downarrow$ & $\Delta_{\mathrm{tgt}}\uparrow$ \\
\midrule
Add crosswalk            & 2{,}771 & 49.6 & 9.2 & 0.071 & $+1{,}772.2$ \\
Delete crosswalk         &    957  & 75.1 & 19.7 & 0.119 & $+4{,}070.4$ \\
Single$\to$double yellow & 1{,}750 & 31.0 & 7.4 & 0.041 &     $+29.8$ \\
\bottomrule
\end{tabular}
\end{table}

Crosswalk deletion is more successful than addition (75.1\% versus 49.6\%). Single-to-double yellow editing achieves 31.0\% success under a road-divider skeleton-length proxy. Deletion also produces the highest false activation (19.7\%), showing that target success does not ensure non-target preservation.
Fixed-seed guidance parameter sweeps on AV2 and Waymo further show that stronger guidance increases both edit success and unintended changes (Tab.~\ref{tab:cfg_sweep}).
These failures are already present in raw outputs before SGR,
indicating limitations in text-to-geometry control.

%% file: text/conclusion.tex
\section{Conclusion}
\label{sec:conclusion}

We introduced \emph{Controllable Road Marking Generation}, a conditional BEV synthesis problem in which the road-marking layer is treated as a generative output rather than as something to be extracted from observations. Given drivable-area geometry, outer-ring context, and a template-derived textual description, our model synthesizes the missing center-region marking layer conditioned on these spatial and semantic inputs.


Our results support conditional road-marking completion with limited semantic control. Evaluation is conducted on models trained on Argoverse~2, Waymo Open Dataset, and a diagnostic nuScenes split, but covers only three marking channels and a narrow external baseline set.
Prompts are template-derived, fine-grained divider editing remains unreliable, and single-reference metrics do not capture all plausible layouts. Future evaluation could use multiple expert-vetted layouts to assess coverage and diversity. Traffic-rule compliance and downstream planning benefits are not validated; generated layouts therefore require independent assessment before practical use.

These limits motivate three extensions:
\begin{itemize}
    \item \textbf{Richer topology.} Link raster synthesis to vector lane graphs, regulatory annotations, and traffic-control elements.
    \item \textbf{Closed-loop testing.} Evaluate generated layouts in closed-loop simulation and planning pipelines to validate traffic-rule compliance and downstream planning benefits.
    \item \textbf{Broader conditioning.} Incorporate design standards, safety priorities, traffic demand, and regional marking vocabularies.
\end{itemize}



%% file: text/appendix.tex
\newpage
\appendix

\section{Semantic prompt specification}
\label{app:semantic_prompt}


This appendix gives the exact templates, vocabulary, and metadata mapping used to automatically generate the natural-language descriptions $\mathbf{m}$ for the CLIP text encoder of Sec.~\ref{ssec:conditioning}.

We generate one deterministic prompt for each tile:
\begin{equation*}
m_i = \mathcal{G}(\text{meta}_i,\; y_i),
\end{equation*}
where $\text{meta}_i$ is the tile metadata JSON and $y_i$ is the lane-marking mask used only for crosswalk counting. Prompts are precomputed offline once and stored as split-specific JSON files. 
Precomputation is preferred over online generation because fixed text removes epoch-to-epoch wording drift and makes training reproducible.

Prompts are precomputed once and stored in \texttt{train\_prompts.json}, \texttt{val\_prompts.json}, and \texttt{test\_prompts.json}, indexed by \texttt{tile\_id}. Training loads these fixed strings directly, avoiding epoch-to-epoch wording changes and online prompt-generation cost.
Each prompt is generated from tile metadata (\texttt{segments}, lane marks, lane types, intersection flag). Only segments overlapping the center $256\times256$ region are used, because supervision and evaluation focus on this region (Sec.~\ref{ssec:data}). Road topology (one-way / two-way), lane count, and divider style are derived by fixed rules from metadata. Crosswalk wording is derived from the pedestrian-crossing mask channel when \texttt{output\_npy} is provided during precomputation; otherwise no crosswalk instance is counted. This keeps text attributes aligned with the mask supervision used by the model.

\paragraph{Template and vocabulary.}
We use a fixed clause order: \texttt{<road topology>, [divider], <center line>, <crosswalk>, [intersection]}. A fixed order reduces linguistic variance and lets the model focus on semantics rather than surface form. Lane-mark terms are canonicalized to short forms (e.g., \texttt{solid white}, \texttt{dashed white}, \texttt{double yellow center line}) to reduce lexical sparsity. Negative clauses (e.g., \texttt{no double yellow}, \texttt{no crosswalks}) are explicit, so absence is also a learnable condition.

\begin{table}[t]
\centering
\small
\setlength{\tabcolsep}{4pt}
\caption{Representative deterministic prompt strings used for text conditioning.}
\label{tab:prompt_examples}
\begin{tabular}{p{0.24\linewidth}p{0.68\linewidth}}
\toprule
\textbf{Semantic case} & \textbf{Example prompt} \\
\midrule
Two-way crossing tile &
``two-way, 2 lanes each direction, solid white dividers, single yellow center line, 4 crosswalks at intersection'' \\
One-way non-crossing tile &
``one-way, 1 lane, no double yellow, no crosswalks'' \\
Double-yellow center line &
``two-way, 2 lanes each direction, solid white dividers, double yellow center line, 1 crosswalk at intersection'' \\
Lane-type context &
``two-way, 2 lanes each direction, solid white dividers, double yellow center line, 2 crosswalks at intersection'' \\
\bottomrule
\end{tabular}
\end{table}

\paragraph{Training-time text settings.}
Text conditioning is enabled with classifier-free dropout~\cite{ho2022classifier} ($\texttt{p\_text\_uncond}{=}0.1$, the value cited in Sec.~\ref{ssec:conditioning}) and inference-time text guidance ($\texttt{text\_cfg\_scale}{=}3.0$). Dropout is applied per sample to expose both conditional and unconditional cases during training, and guidance is applied at sampling to strengthen prompt control.

\paragraph{Controllability parser thresholds.}

Tab.~\ref{tab:control_3ch} of Sec.~\ref{ssec:control} parses each generated mask under per-channel \emph{on}/\emph{off} thresholds. The \emph{on} threshold $\tau^{\mathrm{on}}_{\bullet}$ is set to the median of the GT channel measure over GT-positive tiles (ped-cross area in pixels; lane- and road-divider skeleton length in pixels), giving $\tau^{\mathrm{on}}_{\mathrm{cw}}{=}2274$, $\tau^{\mathrm{on}}_{\mathrm{lane}}{=}195$, and $\tau^{\mathrm{on}}_{\mathrm{road}}{=}243$ on the AV2 test split. The \emph{off} threshold is fixed at $\tau^{\mathrm{off}}_{\bullet}{=}0.10\,\tau^{\mathrm{on}}_{\bullet}$ to admit a tolerance band around true zero before declaring an "off" event. These test-derived thresholds define the diagnostic parser used for Tab.~\ref{tab:control_3ch}. The resulting success rates are calibrated to the empirical AV2 test distribution rather than evaluated with thresholds selected on an independent validation split.

\section{Baseline adaptation details}
\label{app:baseline_adaptation}

This appendix gives implementation details for our adaptation of LSR-DM (Lane Segmentation Refinement with Diffusion Models)~\cite{ruiz2025lane} to the center-region imputation benchmark of Sec.~\ref{ssec:main}. We adopt the LSR-DM UNet, sigmoid variance schedule, v-parameterization, and three-layer aerial-patch encoder, but adapt the inputs, the conditioning signal, and the sampling procedure to fit our task; the resulting ``matched'' baseline shares all inputs available to our main model except text prompts and the blur-then-deblur strategy.

\paragraph{Task adaptation.}
LSR-DM was originally introduced as a Stage-2 lane-mask refinement network: the diffusion model starts from a noised version of a CNN-predicted (Stage-1) lane segmentation mask and denoises it. Our task has no Stage-1 CNN, so this ``conditioned starting latent'' contribution does not apply; we instead start sampling from Gaussian noise and rely on the spatial conditioning and the held-fixed outer-ring observation~$\mathbf{p}$ to anchor the geometry.

\paragraph{Inputs and outputs.}
The baseline receives the same drivable-area mask~$\mathbf{s}$ and partial outer-ring marking observation~$\mathbf{p}$ as our model (Sec.~\ref{ssec:data}), and predicts the missing center markings on the $256{\times}256$ inner window. The 3-channel output (lane dividers, road dividers, pedestrian crossings) and the train / validation / test split are identical to those used by our model, so all numbers in Tab.~\ref{tab:main_structural_3ch} compare on matched data. At every DDIM step, the outer ring is overwritten with the ground-truth observation to keep the boundary condition strict.

\paragraph{Training configuration.}
We retrain LSR-DM from scratch on our training split, at native $512{\times}512$ resolution. Training runs for $180{,}000$ steps with an effective batch size of $16$ ($4$ samples per rank $\times$ $4$~GPUs via DDP) on $4{\times}$ NVIDIA RTX~5090. Optimizer (AdamW, lr $8{\times}10^{-5}$), sigmoid variance schedule (start $=-3$, end $=3$, $\tau=1$), v-parameterization, and 50-step DDIM sampling follow the LSR-DM paper; EMA decay $0.995$ as in the original. Augmentation is restricted to horizontal/vertical flipping; no rotation or color jitter (the latter is moot under a binary drivable-mask conditioning signal).

\paragraph{Architecture notes.}
The original LSR-DM source code is no longer publicly accessible, so we reimplemented the architecture in our framework. The encoder uses linear attention at every resolution; the bottleneck retains standard self-attention; the decoder uses standard self-attention at the lowest resolution and linear attention thereafter. This matches the memory profile required by $512{\times}512$ inputs on a single 32~GB device while staying close to the published description.

\paragraph{Conditioning signal.}
The original method uses an ``aerial patch'' RGB image embedding $e_{\text{context}}$ as a refinement guidance signal. In our setting no aerial imagery is available; we substitute the entire $512{\times}512$ drivable-area mask~$\mathbf{s}$ as the context input, encoded by the same three-layer convolutional aerial-patch encoder of LSR-DM. The partial marking hint~$\mathbf{p}$ enters the model as the outer ring of the denoising target, not through the context encoder. This matches the spatial information available to both models, while differences in text conditioning, architecture, training targets, and post-processing remain part of the comparison.

\section{Gaussian blur parameter selection}
\label{app:gaussian_blur}

The training-target Gaussian blur $\mathcal{B}_\sigma$ and the inference-time SGR structured render are the two halves of the thin-structure stabilization strategy. The full system uses $\sigma{=}2$\,px on the training GT and applies SGR at inference.

\paragraph{Choice of $\sigma$.}
We chose $\sigma{=}2$\,px after qualitative comparison on early prototypes: $\sigma{=}1$ produced visible checkerboarding on dashed lane dividers, $\sigma{=}3$ over-smoothed adjacent parallel dividers into a single blob, and $\sigma{=}2$ retained dash separation while keeping the rectified-flow loss stable through the first $5{,}000$ steps.

\section{Component ablations}
\label{app:components}

We evaluate topology-aware supervision and SGR in a
$2\times2$ ablation on AV2 and Waymo
(Tab.~\ref{tab:topology_sgr}).
Without SGR, raw outputs are skeletonized only for
width-normalized metric computation; no curve fitting
or Gaussian re-rendering is applied.
Enabling both components yields the highest clDice
on both datasets, although Buffered F1 is not
uniformly highest.

\begin{table}[t]
\centering
\small
\setlength{\tabcolsep}{6pt}
\caption{Topology and SGR ablation.
Values are percentages; bold marks the best
result for each metric within each dataset.}
\label{tab:topology_sgr}
\begin{tabular}{lcccc}
\toprule
Dataset & Topology loss & SGR
& Buffered F1$\uparrow$ & clDice$\uparrow$ \\
\midrule
AV2   & No  & No  & 80.5 & 43.4 \\
      & No  & Yes & \textbf{81.8} & 47.2 \\
      & Yes & No  & 78.2 & 48.1 \\
      & Yes & Yes & 76.8 & \textbf{50.2} \\
\midrule
Waymo & No  & No  & \textbf{88.6} & 59.1 \\
      & No  & Yes & 87.3 & 61.9 \\
      & Yes & No  & 86.4 & 62.7 \\
      & Yes & Yes & 88.0 & \textbf{66.2} \\
\bottomrule
\end{tabular}
\end{table}

\section{Evaluation subsets and SGR sensitivity}
\label{app:extended_structural}

Table~6 reports the GT-positive subsets used to compute the per-channel scores in Table~2. Each channel is evaluated on its own subset, determined solely by ground-truth presence. Overall scores are the unweighted mean of the three per-channel scores.

\begin{table*}[t]
\centering
\small
\caption{GT-positive subsets used for per-channel AV2 evaluation. Percentages are relative to the full set of 4,597 test tiles.}
\label{tab:full_per_channel_appendix}
\begin{tabular}{lrr}
\toprule
Channel & GT-positive tiles & Share of test tiles \\
\midrule
Lane divider        & 2,868 & 62.4\% \\
Road divider        & 2,839 & 61.8\% \\
Pedestrian crossing & 1,928 & 41.9\% \\
\bottomrule
\end{tabular}
\end{table*}

\paragraph{SGR parameter sensitivity.}
Across AV2 and Waymo, local sweeps of the threshold, sampling spacing, linking, and angle settings were less sensitive than changes in stroke width. Both validation sets independently selected a stroke width of $1.25$\,px.



\section{Compute resources}
\label{app:compute}

All training and evaluation runs reported in this paper use a single workstation node with $4{\times}$ NVIDIA RTX~5090 GPUs ($32$~GB each), PyTorch~2.5.1, and CUDA~12.4 under DistributedDataParallel. A 180k-step training run for the main model or any of its 180k-step ablations takes roughly one day of wall-clock time on this node. The CNN-deblur ablation trains on a single GPU in a comparable amount of time. Inference on the $4{,}597$-tile AV2 test split takes a few hours per checkpoint with $50$-step sampling; SGR post-processing adds a negligible CPU cost.

Counting all checkpoints that contribute numbers to the paper, the reported results consume on the order of $200$~GPU-hours on this hardware. Earlier development, including dropped architectures and abandoned design directions that do not appear in the paper, used several times this amount of compute on the same node. Earlier development also used A100 nodes; all reported numbers are reproduced on the $4{\times}$ RTX~5090 setup.

\section{Licenses of existing assets}
\label{app:licenses}

Tab.~\ref{tab:licenses} lists the existing datasets, model weights, and software libraries used in this work, together with their licenses and the version or release used. Our research use is non-commercial and complies with the Argoverse~2 CC~BY-NC-SA~4.0 terms; baseline adaptations are described in App.~\ref{app:baseline_adaptation}.
This work also uses the Waymo Open Dataset provided by Waymo LLC under its non-commercial dataset agreement (\url{https://waymo.com/open/terms/}) and nuScenes under its data-use terms (\url{https://www.nuscenes.org/terms-of-use}).

\begin{table}[!htbp]
\centering
\small
\caption{Existing assets, versions, and licenses.}
\label{tab:licenses}
\begin{tabular}{lll}
\toprule
Asset & Version & License \\
\midrule
Argoverse~2 Sensor / Map Dataset & v2.0 & CC BY-NC-SA 4.0 \\
CLIP text encoder & ViT-L/14 (OpenAI) & MIT \\
PyTorch & 2.5.1 & BSD-3-Clause \\
diffusers (HuggingFace) & 0.30.x & Apache-2.0 \\
xFormers & 0.0.28 & BSD-3-Clause \\
NumPy / SciPy / scikit-image & current & BSD-3-Clause \\
\bottomrule
\end{tabular}
\end{table}

The public repository includes an MIT license file. Upstream datasets and pretrained weights remain subject to their respective licenses and terms of use.

\section{Additional qualitative results}
\label{app:more_results}

\begin{figure*}[!htbp]
  \centering
  \includegraphics[width=\textwidth]{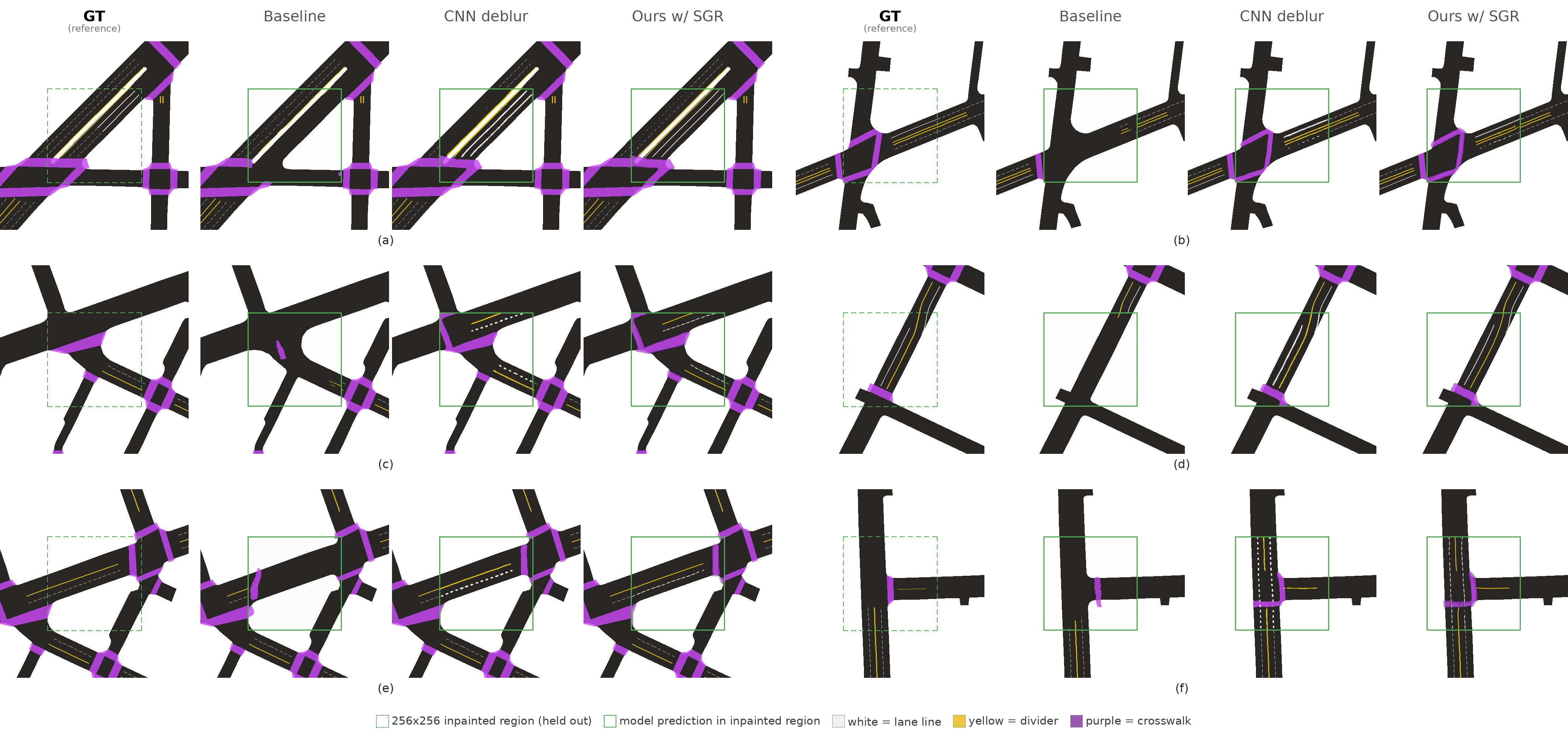}
  \caption{Qualitative comparisons on Argoverse 2 BEV tiles. Each four-column group shows, from left to right, the ground truth, the adapted LSR-DM baseline, our CNN-deblur variant, and our SGR variant. White, yellow, and purple indicate lane dividers, road dividers, and pedestrian crossings, respectively. Green boxes mark the central $256\times256$ target region.}
  \label{fig:qual-appendix}
\end{figure*}

\paragraph{Text-editing failures.}
Representative failures observed before SGR include a short or disconnected second divider stroke, unintended connections at complex intersections, and changes to non-target dividers. Increasing guidance improves edit success but also increases unintended changes (Tab.~\ref{tab:cfg_sweep}).

\begin{table}[t]
\centering
\small
\setlength{\tabcolsep}{5pt}
\caption{
Fixed-seed classifier-free guidance sweep on 300 eligible tiles
per dataset, evaluated on raw outputs before SGR.
Success and false activation (FA) are percentages.
$\Delta_{\mathrm{tgt}}$ denotes direction-aligned target change.
}
\label{tab:cfg_sweep}
\begin{tabular}{llrrrr}
\toprule
Dataset & CFG & Success$\uparrow$ & FA$\downarrow$
& fa-MSE$\downarrow$ & $\Delta_{\mathrm{tgt}}\uparrow$ \\
\midrule
AV2 & 1 & 7.7  & 6.0  & 0.0170 & $+26.8$ \\
    & 3 & 24.3 & 12.7 & 0.0406 & $+66.1$ \\
    & 5 & 29.3 & 17.3 & 0.0707 & $+74.1$ \\
    & 7 & 35.3 & 20.7 & 0.0812 & $+98.5$ \\
\midrule
Waymo & 1 & 12.1 & 4.4  & 0.0110 & $+45.9$ \\
      & 3 & 28.7 & 7.4  & 0.0369 & $+71.9$ \\
      & 5 & 33.4 & 11.5 & 0.0611 & $+84.0$ \\
      & 7 & 46.1 & 18.6 & 0.0698 & $+119.9$ \\
\bottomrule
\end{tabular}
\end{table}

\begin{table*}[t]
\centering
\small
\setlength{\tabcolsep}{3.5pt}
\caption{
Within-dataset structural metrics for independently trained Ours w/ SGR deblur models, evaluated on the generated center $256{\times}256$ region. Results are reported on $261$ nuScenes-SG diagnostic tiles, $4{,}597$ AV2 test tiles, and $1{,}000$ Waymo test tiles. Per-channel scores use GT-positive tiles. Values are percentages.
}
\label{tab:cross_dataset_structural_3ch}
\begin{tabular}{lcccccccc}
\toprule
\multirow{2}{*}{Dataset}
& \multicolumn{2}{c}{Overall}
& \multicolumn{2}{c}{Lane Div.}
& \multicolumn{2}{c}{Road Div.}
& \multicolumn{2}{c}{Ped. Cross.} \\
\cmidrule(lr){2-3}\cmidrule(lr){4-5}\cmidrule(lr){6-7}\cmidrule(lr){8-9}
& F1$\uparrow$ & clDice$\uparrow$
& F1$\uparrow$ & clDice$\uparrow$
& F1$\uparrow$ & clDice$\uparrow$
& F1$\uparrow$ & clDice$\uparrow$ \\
\midrule
nuScenes-SG\textsuperscript{\dag}
& 30.4 & 8.1
& 40.7 & 4.6
& 33.1 & 8.7
& 17.4 & 11.0 \\
AV2
& 76.8 & 50.2
& 74.0 & 28.4
& 66.6 & 38.0
& 89.7 & 84.1 \\
Waymo
& 88.0 & 66.2
& 89.5 & 55.9
& 85.7 & 56.5
& 88.9 & 86.3 \\
\bottomrule
\end{tabular}
\par\smallskip
\footnotesize
\textsuperscript{\dag}nuScenes-SG denotes the Singapore Holland Village diagnostic split.
\end{table*}

%% file: bib/references.bib
@misc{mdot_national_firsts,
  author       = {{Michigan Department of Transportation}},
  title        = {Transportation National Firsts},
  howpublished = {\url{https://www.michigan.gov/mdot/about/history/transportation-national-firsts}},
  note         = {Accessed: 2026-03-08},
  year         = {n.d.}
}

@article{heipke1997evaluation,
  title={Evaluation of automatic road extraction},
  author={Heipke, Christian and Mayer, Helmut and Wiedemann, Christian and Jamet, Olivier},
  journal={International Archives of Photogrammetry and Remote Sensing},
  volume={32},
  number={3 SECT 4W2},
  pages={151--160},
  year={1997},
  publisher={INTERNATIONAL SOCIETY FOR PHOTOGRAMMETRY \& REMOTE}
}

@inproceedings{parish2001procedural,
  title={Procedural modeling of cities},
  author={Parish, Yoav IH and M{\"u}ller, Pascal},
  booktitle={Proceedings of the 28th annual conference on Computer graphics and interactive techniques},
  pages={301--308},
  year={2001}
}

@article{carlson2009benefits,
  title={Benefits of pavement markings: A renewed perspective based on recent and ongoing research},
  author={Carlson, Paul J and Park, Eun Sug and Andersen, Carl K},
  journal={Transportation research record},
  volume={2107},
  number={1},
  pages={59--68},
  year={2009},
  publisher={SAGE Publications Sage CA: Los Angeles, CA}
}

@inproceedings{csurka2013good,
  title={What is a good evaluation measure for semantic segmentation?.},
  author={Csurka, Gabriela and Larlus, Diane and Perronnin, Florent},
  booktitle={Bmvc},
  volume={27},
  number={2013},
  pages={10--5244},
  year={2013},
  organization={Bristol}
}

@inproceedings{mosbock2016importance,
  title={The importance of road markings for road safety and modern traffic management},
  author={Mosbock, H and Burghardt, Tomasz E},
  booktitle={Proceedings of the 1st European Road Infrastructure Congress},
  pages={1--12},
  year={2016}
}

@article{gitelman2017examination,
  title={An examination of the influence of crosswalk marking removal on pedestrian safety as reflected in road user behaviours},
  author={Gitelman, Victoria and Carmel, Roby and Pesahov, Fany and Hakkert, Shalom},
  journal={Transportation research part F: traffic psychology and behaviour},
  volume={46},
  pages={342--355},
  year={2017},
  publisher={Elsevier}
}

@inproceedings{dosovitskiy2017carla,
  title={CARLA: An open urban driving simulator},
  author={Dosovitskiy, Alexey and Ros, German and Codevilla, Felipe and Lopez, Antonio and Koltun, Vladlen},
  booktitle={Conference on robot learning},
  pages={1--16},
  year={2017},
  organization={PMLR}
}

@inproceedings{lin2017focal,
  title={Focal loss for dense object detection},
  author={Lin, Tsung-Yi and Goyal, Priya and Girshick, Ross and He, Kaiming and Doll{\'a}r, Piotr},
  booktitle={Proceedings of the IEEE international conference on computer vision},
  pages={2980--2988},
  year={2017}
}

@inProceedings{Mosinska_2018_CVPR,
author = {Mosinska, Agata and Márquez-Neila, Pablo and Koziński, Mateusz and Fua, Pascal},
title = {Beyond the Pixel-Wise Loss for Topology-Aware Delineation},
booktitle = {Proceedings of the IEEE Conference on Computer Vision and Pattern Recognition (CVPR)},
month = {June},
year = {2018}
}

@article{miao2018computational,
  title={Computational urban design prototyping: Interactive planning synthesis methods—a case study in Cape Town},
  author={Miao, Yufan and Koenig, Reinhard and Knecht, Katja and Konieva, Kateryna and Bu{\v{s}}, Peter and Chang, Mei-Chih},
  journal={International Journal of Architectural Computing},
  volume={16},
  number={3},
  pages={212--226},
  year={2018},
  publisher={SAGE Publications Sage UK: London, England}
}

@article{babic2020road,
  title={Road markings and their impact on driver behaviour and road safety: A systematic review of current findings},
  author={Babi{\'c}, Dario and Fioli{\'c}, Mario and Babi{\'c}, Darko and Gates, Timothy},
  journal={Journal of advanced transportation},
  volume={2020},
  number={1},
  pages={7843743},
  year={2020},
  publisher={Wiley Online Library}
}

@article{li2020differentiable,
  title={Differentiable vector graphics rasterization for editing and learning},
  author={Li, Tzu-Mao and Luk{\'a}{\v{c}}, Michal and Gharbi, Micha{\"e}l and Ragan-Kelley, Jonathan},
  journal={ACM Transactions on Graphics (TOG)},
  volume={39},
  number={6},
  pages={1--15},
  year={2020},
  publisher={ACM New York, NY, USA}
}

@inproceedings{shit2021cldice,
  title={clDice-a novel topology-preserving loss function for tubular structure segmentation},
  author={Shit, Suprosanna and Paetzold, Johannes C and Sekuboyina, Anjany and Ezhov, Ivan and Unger, Alexander and Zhylka, Andrey and Pluim, Josien PW and Bauer, Ulrich and Menze, Bjoern H},
  booktitle={Proceedings of the IEEE/CVF conference on computer vision and pattern recognition},
  pages={16560--16569},
  year={2021}
}

@article{zhang2021deep,
  title={Deep learning in lane marking detection: A survey},
  author={Zhang, Youcheng and Lu, Zongqing and Zhang, Xuechen and Xue, Jing-Hao and Liao, Qingmin},
  journal={IEEE Transactions on Intelligent Transportation Systems},
  volume={23},
  number={7},
  pages={5976--5992},
  year={2021},
  publisher={IEEE}
}

@inproceedings{tan2021scenegen,
  title={Scenegen: Learning to generate realistic traffic scenes},
  author={Tan, Shuhan and Wong, Kelvin and Wang, Shenlong and Manivasagam, Sivabalan and Ren, Mengye and Urtasun, Raquel},
  booktitle={Proceedings of the IEEE/CVF Conference on Computer Vision and Pattern Recognition},
  pages={892--901},
  year={2021}
}

@inproceedings{radford2021learning,
  title={Learning transferable visual models from natural language supervision},
  author={Radford, Alec and Kim, Jong Wook and Hallacy, Chris and Ramesh, Aditya and Goh, Gabriel and Agarwal, Sandhini and Sastry, Girish and Askell, Amanda and Mishkin, Pamela and Clark, Jack and others},
  booktitle={International conference on machine learning},
  pages={8748--8763},
  year={2021},
  organization={PMLR}
}

@article{ho2022classifier,
  title={Classifier-free diffusion guidance},
  author={Ho, Jonathan and Salimans, Tim},
  journal={arXiv preprint arXiv:2207.12598},
  year={2022}
}

@inproceedings{rombach2022high,
  title={High-resolution image synthesis with latent diffusion models},
  author={Rombach, Robin and Blattmann, Andreas and Lorenz, Dominik and Esser, Patrick and Ommer, Bj{\"o}rn},
  booktitle={Proceedings of the IEEE/CVF conference on computer vision and pattern recognition},
  pages={10684--10695},
  year={2022}
}

@article{liao2022maptr,
  title={Maptr: Structured modeling and learning for online vectorized hd map construction},
  author={Liao, Bencheng and Chen, Shaoyu and Wang, Xinggang and Cheng, Tianheng and Zhang, Qian and Liu, Wenyu and Huang, Chang},
  journal={arXiv preprint arXiv:2208.14437},
  year={2022}
}

@inproceedings{he2022lane,
  title={Lane-level street map extraction from aerial imagery},
  author={He, Songtao and Balakrishnan, Hari},
  booktitle={Proceedings of the IEEE/CVF Winter Conference on Applications of Computer Vision},
  pages={2080--2089},
  year={2022}
}

@inproceedings{feng2022rethinking,
  title={Rethinking efficient lane detection via curve modeling},
  author={Feng, Zhengyang and Guo, Shaohua and Tan, Xin and Xu, Ke and Wang, Min and Ma, Lizhuang},
  booktitle={Proceedings of the IEEE/CVF conference on computer vision and pattern recognition},
  pages={17062--17070},
  year={2022}
}

@inproceedings{ma2022towards,
  title={Towards layer-wise image vectorization},
  author={Ma, Xu and Zhou, Yuqian and Xu, Xingqian and Sun, Bin and Filev, Valerii and Orlov, Nikita and Fu, Yun and Shi, Humphrey},
  booktitle={Proceedings of the IEEE/CVF Conference on Computer Vision and Pattern Recognition},
  pages={16314--16323},
  year={2022}
}

@inproceedings{liu2022flow,
  title={Flow straight and fast: Learning to generate and transfer data with rectified flow},
  author={Liu, Xingchao and Gong, Chengyue and Liu, Qiang},
  booktitle={The Eleventh International Conference on Learning Representations},
  year={2023}
}

@inproceedings{lugmayr2022repaint,
  title={Repaint: Inpainting using denoising diffusion probabilistic models},
  author={Lugmayr, Andreas and Danelljan, Martin and Romero, Andres and Yu, Fisher and Timofte, Radu and Van Gool, Luc},
  booktitle={Proceedings of the IEEE/CVF conference on computer vision and pattern recognition},
  pages={11461--11471},
  year={2022}
}

@inproceedings{alayrac2022flamingo,
author = {Alayrac, Jean-Baptiste and Donahue, Jeff and Luc, Pauline and Miech, Antoine and Barr, Iain and Hasson, Yana and Lenc, Karel and Mensch, Arthur and Millican, Katherine and Reynolds, Malcolm and Ring, Roman and Rutherford, Eliza and Cabi, Serkan and Han, Tengda and Gong, Zhitao and Samangooei, Sina and Monteiro, Marianne and Menick, Jacob L and Borgeaud, Sebastian and Brock, Andy and Nematzadeh, Aida and Sharifzadeh, Sahand and Bi\'{n}kowski, Miko\l aj and Barreira, Ricardo and Vinyals, Oriol and Zisserman, Andrew and Simonyan, Kar\'{e}n},
booktitle = {Advances in Neural Information Processing Systems},
doi = {10.52202/068431-1723},
editor = {S. Koyejo and S. Mohamed and A. Agarwal and D. Belgrave and K. Cho and A. Oh},
pages = {23716--23736},
publisher = {Curran Associates, Inc.},
title = {Flamingo: a Visual Language Model for Few-Shot Learning},
volume = {35},
year = {2022}
}

@article{pronovost2023scenario,
  title={Scenario diffusion: Controllable driving scenario generation with diffusion},
  author={Pronovost, Ethan and Ganesina, Meghana Reddy and Hendy, Noureldin and Wang, Zeyu and Morales, Andres and Wang, Kai and Roy, Nick},
  journal={Advances in Neural Information Processing Systems},
  volume={36},
  pages={68873--68894},
  year={2023}
}

@inproceedings{liu2023vectormapnet,
  title={Vectormapnet: End-to-end vectorized hd map learning},
  author={Liu, Yicheng and Yuan, Tianyuan and Wang, Yue and Wang, Yilun and Zhao, Hang},
  booktitle={International Conference on Machine Learning},
  pages={22352--22369},
  year={2023},
  organization={PMLR}
}

@article{chen2023polydiffuse,
  title={Polydiffuse: Polygonal shape reconstruction via guided set diffusion models},
  author={Chen, Jiacheng and Deng, Ruizhi and Furukawa, Yasutaka},
  journal={Advances in neural information processing systems},
  volume={36},
  pages={1863--1888},
  year={2023}
}

@inproceedings{peebles2023scalable,
  title={Scalable diffusion models with transformers},
  author={Peebles, William and Xie, Saining},
  booktitle={Proceedings of the IEEE/CVF international conference on computer vision},
  pages={4195--4205},
  year={2023}
}

@inproceedings{li2023gligen,
  title={Gligen: Open-set grounded text-to-image generation},
  author={Li, Yuheng and Liu, Haotian and Wu, Qingyang and Mu, Fangzhou and Yang, Jianwei and Gao, Jianfeng and Li, Chunyuan and Lee, Yong Jae},
  booktitle={Proceedings of the IEEE/CVF conference on computer vision and pattern recognition},
  pages={22511--22521},
  year={2023}
}

@inproceedings{qiao2023end,
  title={End-to-end vectorized hd-map construction with piecewise bezier curve},
  author={Qiao, Limeng and Ding, Wenjie and Qiu, Xi and Zhang, Chi},
  booktitle={Proceedings of the IEEE/CVF Conference on Computer Vision and Pattern Recognition},
  pages={13218--13228},
  year={2023}
}

@inproceedings{wang2023segrefiner,
 author = {Wang, Mengyu and Ding, Henghui and Liew, Jun Hao and Liu, Jiajun and Zhao, Yao and Wei, Yunchao},
 booktitle = {Advances in Neural Information Processing Systems},
 editor = {A. Oh and T. Naumann and A. Globerson and K. Saenko and M. Hardt and S. Levine},
 pages = {79761--79780},
 publisher = {Curran Associates, Inc.},
 title = {SegRefiner: Towards Model-Agnostic Segmentation Refinement with Discrete Diffusion Process},
 url = {https://proceedings.neurips.cc/paper_files/paper/2023/file/fc0cc55dca3d791c4a0bb2d8ddeefe4f-Paper-Conference.pdf},
 volume = {36},
 year = {2023}
}

@article{wilson2023argoverse,
  title={Argoverse 2: Next generation datasets for self-driving perception and forecasting},
  author={Wilson, Benjamin and Qi, William and Agarwal, Tanmay and Lambert, John and Singh, Jagjeet and Khandelwal, Siddhesh and Pan, Bowen and Kumar, Ratnesh and Hartnett, Andrew and Pontes, Jhony Kaesemodel and others},
  journal={arXiv preprint arXiv:2301.00493},
  year={2023}
}

@inproceedings{yuan2024streammapnet,
  title={Streammapnet: Streaming mapping network for vectorized online hd map construction},
  author={Yuan, Tianyuan and Liu, Yicheng and Wang, Yue and Wang, Yilun and Zhao, Hang},
  booktitle={Proceedings of the IEEE/CVF Winter Conference on Applications of Computer Vision},
  pages={7356--7365},
  year={2024}
}

@inproceedings{huang20242d,
  title={2d gaussian splatting for geometrically accurate radiance fields},
  author={Huang, Binbin and Yu, Zehao and Chen, Anpei and Geiger, Andreas and Gao, Shenghua},
  booktitle={ACM SIGGRAPH 2024 conference papers},
  pages={1--11},
  year={2024}
}

@inproceedings{zhang2024gaussianimage,
  title={Gaussianimage: 1000 fps image representation and compression by 2d gaussian splatting},
  author={Zhang, Xinjie and Ge, Xingtong and Xu, Tongda and He, Dailan and Wang, Yan and Qin, Hongwei and Lu, Guo and Geng, Jing and Zhang, Jun},
  booktitle={European Conference on Computer Vision},
  pages={327--345},
  year={2024},
  organization={Springer}
}

@article{zhou2024diffgs,
  title={Diffgs: Functional gaussian splatting diffusion},
  author={Zhou, Junsheng and Zhang, Weiqi and Liu, Yu-Shen},
  journal={Advances in Neural Information Processing Systems},
  volume={37},
  pages={37535--37560},
  year={2024}
}

@article{liu2024enhancing,
  title={Enhancing boundary segmentation for topological accuracy with skeleton-based methods},
  author={Liu, Chuni and Ma, Boyuan and Ban, Xiaojuan and Xie, Yujie and Wang, Hao and Xue, Weihua and Ma, Jingchao and Xu, Ke},
  journal={arXiv preprint arXiv:2404.18539},
  year={2024}
}

@inproceedings{monninger2025mapdiffusion,
  title={MapDiffusion: Generative diffusion for vectorized online HD map construction and uncertainty estimation in autonomous driving},
  author={Monninger, Thomas and Zhang, Zihan and Mo, Zhipeng and Anwar, Md Zafar and Staab, Steffen and Ding, Sihao},
  booktitle={2025 IEEE/RSJ International Conference on Intelligent Robots and Systems (IROS)},
  pages={4099--4106},
  year={2025},
  organization={IEEE}
}

@article{biermeier2025road,
  title={Road marking visibility for automated vehicles: Machine detectability and maintenance standards},
  author={Biermeier, Stefan and Kemper, Dirk and Garc{\'\i}a-Hernandez, Alvaro},
  journal={Case Studies in Construction Materials},
  volume={22},
  pages={e04430},
  year={2025},
  publisher={Elsevier}
}

@article{agrawal2025loid,
  title={Loid: lane occlusion inpainting and detection for enhanced autonomous driving systems},
  author={Agrawal, Aayush and Sivakumar, Ashmitha Jaysi and Kaif, Ibrahim and Banerjee, Chayan},
  journal={Machine Vision and Applications},
  volume={36},
  number={6},
  pages={120},
  year={2025},
  publisher={Springer}
}

@article{liu2025bezier,
  title={Bezier Splatting for Fast and Differentiable Vector Graphics Rendering},
  author={Liu, Xi and Zhou, Chaoyi and Zhao, Nanxuan and Huang, Siyu},
  journal={The Thirty-ninth Annual Conference on Neural Information Processing Systems},
  year={2025}
}

@inproceedings{gao2025curve,
  title={Curve-Aware Gaussian Splatting for 3D Parametric Curve Reconstruction},
  author={Gao, Zhirui and Yi, Renjiao and Dai, Yaqiao and Zhu, Xuening and Chen, Wei and Zhu, Chenyang and Xu, Kai},
  booktitle={Proceedings of the IEEE/CVF International Conference on Computer Vision},
  pages={27531--27541},
  year={2025}
}

@InProceedings{ying2025sketchsplat,
    author    = {Ying, Haiyang and Zwicker, Matthias},
    title     = {SketchSplat: 3D Edge Reconstruction via Differentiable Multi-view Sketch Splatting},
    booktitle = {Proceedings of the IEEE/CVF International Conference on Computer Vision (ICCV)},
    month     = {October},
    year      = {2025},
    pages     = {25649-25659}
}

@article{ruiz2025lane,
  title={Lane Graph Extraction from Aerial Imagery via Lane Segmentation Refinement with Diffusion Models},
  author={Ruiz, Antonio and Melnik, Andrew and Savioli, Nicolo and Wang, Dong and Zhang, Yanfeng and Ritter, Helge},
  journal={Remote Sensing},
  volume={17},
  number={16},
  pages={2845},
  year={2025},
  publisher={MDPI}
}

@inproceedings{wang2025lanediffusion,
  title={LaneDiffusion: Improving Centerline Graph Learning via Prior Injected BEV Feature Generation},
  author={Wang, Zijie and Zhang, Weiming and Zhang, Wei and Tan, Xiao and Liu, Hongxing and Wang, Yaowei and Li, Guanbin},
  booktitle={Proceedings of the IEEE/CVF International Conference on Computer Vision},
  pages={27052--27062},
  year={2025}
}

@article{cira2026roadmark,
  title={RoadMark-cGAN: Generative Conditional Learning to Directly Map Road Marking Lines from Aerial Orthophotos via Image-to-Image Translation},
  author={Cira, Calimanut-Ionut and Yokoya, Naoto and Manso-Callejo, Miguel-{\'A}ngel and Alcarria, Ramon and Broni-Bediako, Clifford and Xia, Junshi and Bordel, Borja},
  journal={Electronics},
  volume={15},
  number={1},
  pages={224},
  year={2026},
  publisher={MDPI}
}

@inproceedings{sun2020waymo,
  title={Scalability in Perception for Autonomous Driving: {Waymo Open Dataset}},
  author={Sun, Pei and Kretzschmar, Henrik and Dotiwalla, Xerxes and Chouard, Aurelien and Patnaik, Vijaysai and Tsui, Paul and Guo, James and Zhou, Yin and Chai, Yuning and Caine, Benjamin and Vasudevan, Vijay and Han, Wei and Ngiam, Jiquan and Zhao, Hang and Timofeev, Aleksei and Ettinger, Scott and Krivokon, Maxim and Gao, Amy and Joshi, Aditya and Zhang, Yu and Shlens, Jonathon and Chen, Zhifeng and Anguelov, Dragomir},
  booktitle={Proceedings of the IEEE/CVF Conference on Computer Vision and Pattern Recognition},
  pages={2446--2454},
  year={2020}
}

@inproceedings{caesar2020nuscenes,
  title={nuscenes: A multimodal dataset for autonomous driving},
  author={Caesar, Holger and Bankiti, Varun and Lang, Alex H and Vora, Sourabh and Liong, Venice Erin and Xu, Qiang and Krishnan, Anush and Pan, Yu and Baldan, Giancarlo and Beijbom, Oscar},
  booktitle={2020 IEEE/CVF conference on computer vision and pattern recognition (CVPR)},
  pages={11618--11628},
  year={2020},
  organization={IEEE}
}

@inproceedings{brandes2026mapdreamer,
  title={MapDreamer: Aerial Imagery Conditioned Latent Diffusion for Lane-Level Map Generation},
  author={Brandes, Julian and Crocoll, Philipp and Burgard, Wolfram},
  booktitle={European Conference on Computer Vision},
  pages={94--111},
  year={2026},
  organization={Springer}
}

@article{farag2026controlmap,
  title={ControlMap: Controllable High-Definition Map Generation for Traffic Scenario Simulation},
  author={Farag, Marwan and W{\"a}ldele, Steffen and Yao, Yu},
  journal={arXiv preprint arXiv:2606.15930},
  year={2026}
}
